\documentclass{article} 
\usepackage[preprint]{colm2026_conference}

\usepackage{microtype}
\usepackage{hyperref}
\usepackage{url}
\usepackage{booktabs}
\usepackage[colorinlistoftodos,prependcaption,textsize=tiny]{todonotes}
\usepackage{subcaption}
\usepackage{xspace}
\usepackage{wrapfig}

\usepackage[most]{tcolorbox}
\usepackage{xcolor}

\definecolor{PromptBlue}{RGB}{28,73,163}
\definecolor{PromptRed}{RGB}{176,25,65}

\newtcolorbox{promptblock}{
  enhanced,
  breakable,
  colback=gray!10,
  colframe=gray!55,
  boxrule=0.5pt,
  arc=2.5mm,
  left=1.4mm,
  right=1.4mm,
  top=1.0mm,
  bottom=1.0mm
}

\usepackage{lineno}

\definecolor{darkblue}{rgb}{0, 0, 0.5}
\hypersetup{colorlinks=true, citecolor=darkblue, linkcolor=darkblue, urlcolor=darkblue}

\title{\methodname: Building Self-Evolving Agent Skill Libraries from Heterogeneous Scientific Resources}

{
\author{%
  \textbf{Shuaike Shen}$^{1, *}$, \textbf{Wenduo Cheng}$^{1, *}$, \textbf{Mingqian Ma}$^2$, \textbf{Alistair Turcan}$^1$,\\\textbf{Martin Jinye Zhang}$^1$, \textbf{Jian Ma}$^{1\dagger}$\\ 
$^1$Ray and Stephanie Lane Computational Biology Department,\\ School of Computer Science, Carnegie Mellon University
\\$^2$Machine Learning Department,\\ School of Computer Science, Carnegie Mellon University\\
$^*$Equal contribution, $^\dagger$Correspondence: \texttt{jianma@cs.cmu.edu}\\
}
}

\begin{document}

\ifcolmsubmission
\linenumbers
\fi

\maketitle

\begin{abstract}
Modern scientific ecosystems are rich in procedural knowledge across repositories, APIs, scripts, notebooks, documentation, databases, and papers, yet much of this knowledge remains fragmented across heterogeneous artifacts that agents cannot readily operationalize. 
This gap between abundant scientific know-how and usable agent capabilities is a key bottleneck for building effective scientific agents.
We present \methodname, a self-evolving framework that converts such resources into validated agent skills, reusable packages that encode task scope, inputs and outputs, execution steps, environment assumptions, provenance, and tests.
\methodname organizes a target domain as a domain knowledge tree, mines resources from high-value branches, extracts operational contracts, compiles them into executable skill packages, and then iteratively expands, repairs, merges, or prunes the resulting library through a closed-loop validation process.
\methodname~produces a substantially novel and internally valid skill library, with 71.1\% of mined skills differing from existing skill libraries such as SkillHub and SkillSMP. We demonstrate that these mined skills improve coding agent performance on five of the six MoSciBench datasets. We further show that \methodname~can design new task-specific skills on demand for concrete scientific objectives, and that the resulting skills substantially improve performance on two challenging genomics tasks: cell type annotation and the scDRS workflow.
Together, these results show that automatically mined skills improve agent performance on benchmarks and domain-specific tasks, expand coverage beyond hand-crafted skill libraries, 
and provide a practical foundation for more capable scientific agents.
Code and an interactive website dashboard for exploring the skill library, domain knowledge tree, and mined resources are available at \url{https://github.com/ma-compbio-lab/SkillFoundry} and \url{https://ma-compbio-lab.github.io/SkillFoundry/}.
\end{abstract}

\section{Introduction}

LLM agents have demonstrated strong capabilities in reasoning, planning, tool use, and interacting with external environments across a wide range of applications~\citep{wang2024survey}. 
However, they often lack the contextual knowledge and domain expertise needed to perform real-world tasks reliably~\citep{xu2026agent}, especially in scientific settings where reliable execution depends not only on access to tools, but also on procedural know-how: 
specialized software conventions, environment-specific constraints, implicit assumptions, and complex multi-step workflows that are rarely available in a form agents can directly use.

Agent skills provide a natural abstraction for bridging this gap.
A skill packages reusable procedural knowledge into a structured, executable unit that specifies what a capability does, how it should be invoked, what inputs and outputs it expects, and what dependencies or constraints it requires. 
In modern agent frameworks, skills are represented as self-contained packages that agents can discover, load, and invoke when relevant~\citep{anthropic_agent_skills_overview}.


Unlike raw tools, which mainly expose callable interfaces, skills capture richer procedural guidance for coordinating multi-step decisions and actions, making them a more reliable foundation for workflow execution~\citep{xu2026agent}.
However, most agent skills today are still hand-crafted, narrowly scoped, and evaluated only in limited settings~\citep{alzubi2026evoskill}, making them difficult to scale across diverse scientific domains.
Recent work has demonstrated the potential of LLM agents to accelerate scientific discovery and automate complex workflows~\citep{huang2025biomni, wang2025geneagent, swanson2025virtual, qu2025crispr, wang2025spatialagent}. 
Yet most existing systems are still built on custom wrappers and predefined tools, making them difficult to scale, adapt, and extend across domains~\citep{gao2025democratizing}. 
At the same time, modern scientific ecosystems already contain a wealth of procedural knowledge in repositories, software packages, APIs, databases, scripts, notebooks, documentation, papers, and web services. 
The challenge is therefore not a lack of scientific know-how, but that much of this knowledge is buried in heterogeneous, fragmented, and unstructured resources that agents cannot directly retrieve, execute, validate, or compose into reliable workflows.

In this work, we ask whether scientific procedural knowledge hidden in heterogeneous domain resources can be automatically converted into agent skills that support reliable and scalable scientific agents. 
To this end, we propose \methodname, a self-evolving framework that transforms heterogeneous scientific resources into structured, executable skills. 
The framework mines domain resources, extracts operational contracts, validates executability, and iteratively refines the resulting skill library. 
Beyond building a reusable library, \methodname~can also synthesize and test task-specific skills on demand for a concrete scientific objective.


We evaluate \methodname~from two complementary perspectives: the quality of the skill library it builds, and the downstream utility of those skills on scientific tasks. 
At the library level, we study whether the mined skills are internally valid, executable, and novel relative to existing skill libraries. 
At the task level, we test whether these skills improve agent performance both when drawn from the existing library and when newly constructed for concrete user-specified workflows. 
Concretely, we evaluate the utility of the mined skills on the scientific discovery benchmark MoSciBench, and study task-specific skill design on two challenging genomics workflows: cell type annotation for spatial transcriptomics data and identification of disease-relevant cells using the statistical genetics tool scDRS~\citep{zhang2022polygenic}. Across these settings, \methodname~produces a substantially novel and internally valid skill library, improves benchmark performance on five of six MoSciBench datasets, and substantially boosts agent performance on both cell type annotation and the scDRS workflow.

This paper makes the following contributions:

\begin{itemize}
\item We introduce \methodname, a self-evolving framework that converts heterogeneous domain resources into a library of structured, executable, and validated agent skills.
\item We establish an evaluation protocol that combines library-level validation, including executability and novelty, with task-level utility on benchmarks and realistic scientific workflows.
\item We demonstrate that the skills produced by \methodname~extend beyond existing skill libraries and improve agent performance on scientific benchmarks.
\item We show that \methodname~can also design task-specific skills for real-world scientific workflows.
\end{itemize}

\section{Related Work}

\textbf{Agent Tools and MCPs.} Prior work has improved LLM agents’ ability to access external capabilities through tool-use learning, API calling, and benchmarked tool-augmented agent systems~\citep{schick2023toolformer,patil2024gorilla,li2023api,qin2023toolllm}. However, these lines of work mostly assume the needed capabilities are already available in usable form. Meanwhile, MCP provides a standard way to expose tools, resources, and prompts to agents~\citep{anthropic_mcp_2024}. More broadly, tools, MCPs, and skills are related but distinct components of the agent ecosystem: tools provide executable interfaces, MCP standardizes access to external capabilities, and skills package reusable procedural knowledge for using such capabilities effectively. Our focus is on this skill layer, rather than on tool-use policy or protocol design alone.


\textbf{Agent Skills.} Prior work has shown that agents benefit from reusable skill libraries. Voyager~\citep{wang2023voyager} introduced an ever-growing library of executable skills for lifelong embodied learning, while Anthropic’s Claude framework formalized agent skills as modular packages that combine instructions, metadata, and optional scripts or resources~\citep{anthropic_agent_skills_overview}. More recent work has explored different ways of building and improving such skills, including reinforcement-learning-based skill reuse~\citep{wang2025reinforcement}, large engineered skill bases for computer-use agents~\citep{chen2026cua}, composable reasoning modules~\citep{jiao2026agentic}, and iterative skill discovery through failure analysis~\citep{alzubi2026evoskill}. Together, these studies highlight the value of reusable and evolving skills. Our setting differs in that we focus on mining skills from heterogeneous domain resources, rather than constructing them primarily from trajectories, failures, reinforcement learning, or manual engineering.

\textbf{Tool Ecosystems and Automatic Tool Discovery.}
A closely related line of work focuses on building large tool ecosystems or automatically converting external artifacts into agent-usable tools. ToolUniverse~\citep{gao2025democratizing} presents a large scientific ecosystem for AI scientists, standardizing access to more than 1,000 machine learning models, datasets, APIs, and scientific packages. Deploy-Master~\citep{wang2026deploy} addresses large-scale tool discovery by identifying more than 50,000 runnable tools from public repositories. Recent work has also explored automatic construction from specific artifact types: Paper2Agent~\citep{miao2025paper2agent} converts papers and associated codebases into interactive MCP-based agents, while ToolRosetta~\citep{di2026toolrosetta} translates open-source repositories and APIs into MCP-compatible tools with an emphasis on reliable invocation and security inspection. 
Together, these methods move beyond manually curated tools and substantially expand the action space available to agents. 
However, they remain primarily tool-centric, focusing on exposing executable interfaces or constructing runnable tools and agents, often from a single artifact type at a time. 
In contrast, our work is skill-centric.

\textbf{Scientific Agents and Tool Use.}
Recent work has explored both domain-specific and more general scientific agent systems. Domain-specific agents such as ChemCrow~\citep{bran2023chemcrow}, SpatialAgent~\citep{wang2025spatialagent}, and CRISPR-GPT~\citep{qu2025crispr} demonstrate the value of integrating specialized tools and workflows tailored to chemistry, spatial biology, and CRISPR-based gene editing, respectively. More general systems such as Biomni~\citep{huang2025biomni} aim to support a broader range of biomedical tasks through larger tool ecosystems. These works collectively highlight both the promise and the challenges of scientific tool use, particularly the need for strong domain-specific understanding, since even small errors in tool selection, parameterization, or execution can lead to invalid results or misleading conclusions~\citep{wei2025ai}. Our work is complementary to this line of research: rather than focusing on a particular scientific domain or agent system, we study how reusable procedural skills can be mined, validated, and refined from heterogeneous domain resources to support more capable scientific agents.

\section{Method}
\label{sec:method}
\subsection{Method Overview}

\methodname~is a tree-guided framework for automatically building and maintaining a domain skill library from heterogeneous scientific resources. Its core idea is to use a domain knowledge tree both as a search prior and as the object being updated: under-covered branches trigger targeted resource mining, mined artifacts are compiled into executable skills, and evaluation signals determine whether the tree should be expanded, revised, or pruned. This design turns open-ended skill collection into a closed-loop acquisition process.

\begin{figure}[t]
\centering
\includegraphics[width=1\linewidth]{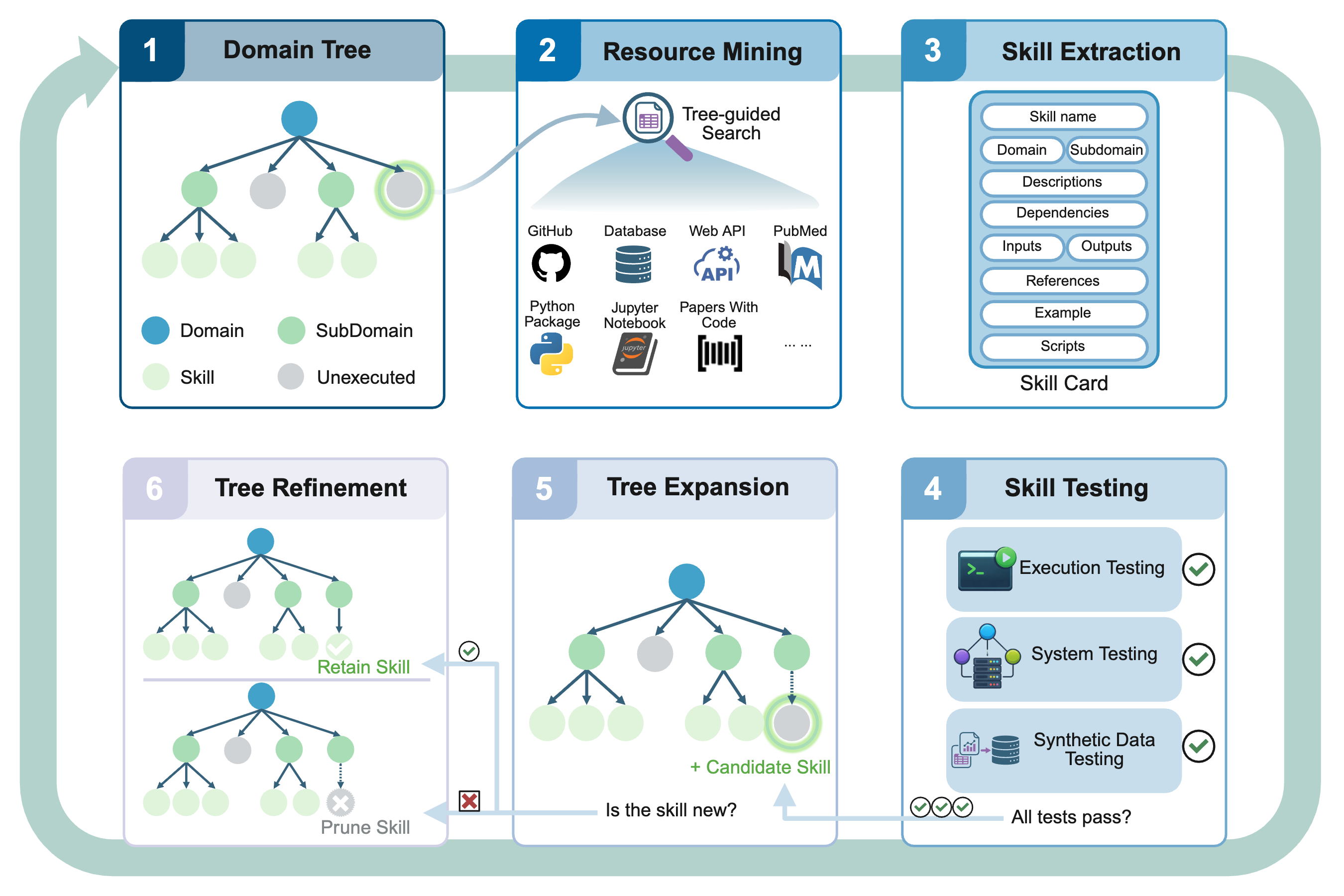}
\vspace{-1em}
\caption{Overview of \methodname. 
Starting from a domain knowledge tree, the framework mines branch-relevant resources, extracts structured candidate skills, validates them through multi-level testing, expands the tree with verified skills, and prunes redundant or low-value leaves.}
\vspace{-1em}
\label{Fig:overview}
\end{figure}

Figure~\ref{Fig:overview} illustrates this loop. Starting from a domain tree (Step 1), \methodname~selects a branch and mines relevant resources such as repositories, databases, APIs, papers, and notebooks (Step 2). The retrieved artifacts are compiled into structured skill cards with explicit scope, dependencies, inputs, outputs, sources, and examples (Step 3). Each skill then undergoes execution testing, system testing, and synthetic-data testing (Step 4). Skills that pass are added as new candidate leaves (Step 5), while redundant or low-value skills are consolidated or removed during tree refinement (Step 6). The updated tree then guides the next round of mining, yielding a self-evolving skill library.

\subsection{Domain Knowledge Tree Construction}

We represent the target field as a rooted tree $T=(V,E)$, where internal nodes denote domains and subdomains, and leaves denote actionable skill targets. Rather than serving as a static taxonomy, this tree acts as the control structure for adaptive mining. Each node stores lightweight state, including linked resources, existing skills, and validation status, so the system can distinguish well-covered regions from high-value frontiers. As a result, \methodname~prioritizes branches with abundant resources but weak verified skill coverage, rather than searching the resource space uniformly.

The tree is initialized from a manually curated but extensible taxonomy and is updated as the library grows. Broad branches can be split when mining reveals stable subareas, while stale or redundant leaves can be merged or pruned. In this way, the tree serves both as a structured prior over the domain and as a compact summary of the current state of library coverage.

\subsection{Tree-Guided Skill Mining}

Given the current tree, \methodname~runs a staged mining loop that selects a focus branch, retrieves relevant resources, and compiles candidate skills. In the implementation, this loop is instantiated as \texttt{tree\_check} $\rightarrow$ \texttt{resource\_search} $\rightarrow$ \texttt{skill\_build} $\rightarrow$ \texttt{skill\_test} $\rightarrow$ \texttt{refresh}. Resource search prioritizes authoritative artifacts, such as official documentation, maintained repositories, package references, workflows, notebooks, and method papers, so that skill induction is grounded in reliable sources rather than generic web text.

From the retrieved artifacts, the system extracts an operational contract and compiles it into a reusable skill package. As reflected by the skill card in Fig.~\ref{Fig:overview}, each skill records its scope, dependencies, inputs, outputs, provenance, and example usage. In the repository, a compiled skill includes both human-readable instructions and machine-readable metadata, together with executable scripts or tests when available. The refresh stage then updates the registry so that subsequent iterations operate on an accurate view of coverage and gaps in the library.

\subsection{Skill Testing and Tree Refinement}

A candidate skill is added to the library only after multi-stage validation. As shown in Step 4 of Fig.~\ref{Fig:overview}, \methodname~applies three complementary tests. Execution testing checks whether the skill runs under its declared contract. System testing validates skills whose realistic use depends on infrastructure such as SLURM. Synthetic-data testing is used when live resources are unavailable, expensive, unstable, or difficult to exercise repeatedly during development. Rather than approximating the full downstream task, it checks whether the skill correctly consumes controlled inputs, supports all required arguments and file interfaces specified by the skill contract, and produces stable outputs under fixed mock conditions. In this sense, it verifies both contract completeness and behavioral stability. Here, contract completeness refers to whether the skill specifies all inputs, outputs, and execution assumptions needed for reliable use, while behavioral stability refers to whether it produces consistent outputs under the same controlled inputs rather than varying with hidden state or incidental changes in the environment. Together, these tests establish a minimum standard of executability and robustness before a skill is accepted into the library.

Evaluation results directly update the tree. Skills that pass are inserted as candidate new leaves and expand coverage in the corresponding branch, as shown in Step 5. Skills that fail repeatedly, duplicate existing functionality, or contribute little marginal value are revised, merged, or pruned during refinement, as shown in Step 6. This final step closes the loop, since \methodname~does not simply accumulate skills, but continuously reorganizes the tree according to empirical signals of correctness, usefulness, and novelty.

\section{Experiments}

\begin{figure}[!t]
\centering
\includegraphics[width=\linewidth]{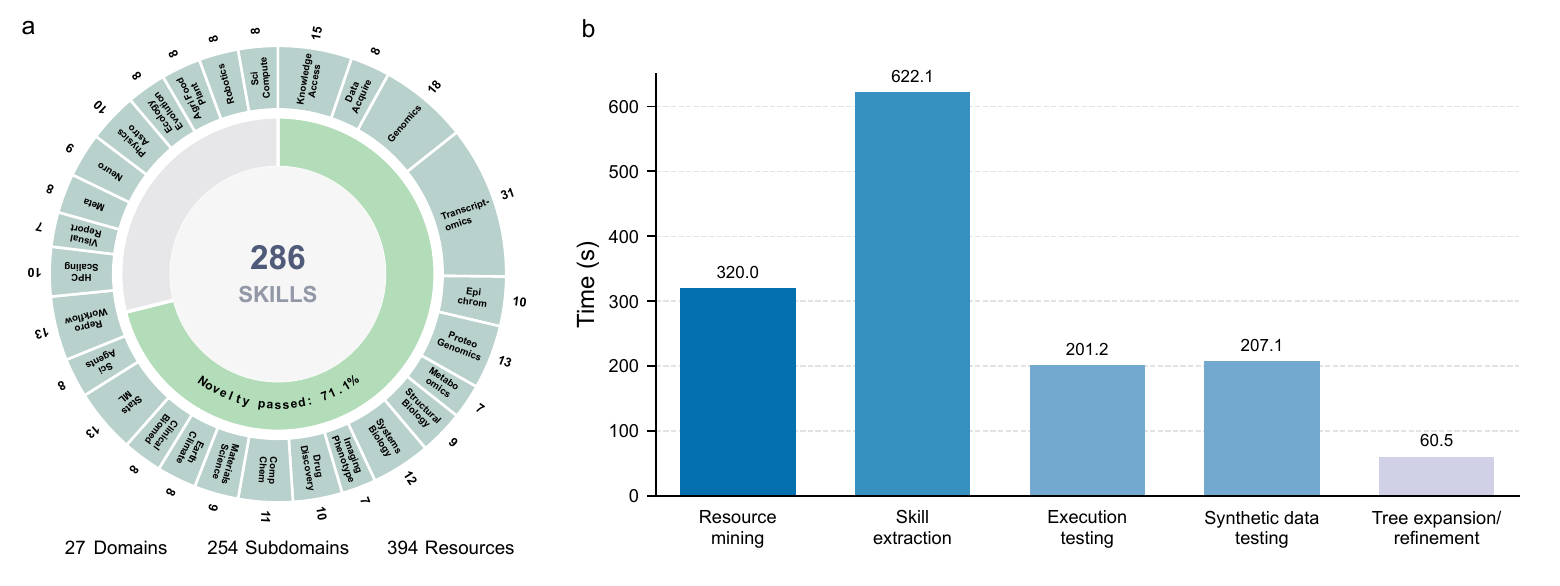}
\vspace{-1.8em}
\caption{
Composition of the mined skill library and runtime of the mining pipeline. 
(a) Distribution of the 286 mined skills across domains and subdomains. 
(b) Average runtime of the major pipeline stages under a fixed resource-mining budget.}
\vspace{-0.8em}
\label{fig:novelty_runtime}
\end{figure}

Having introduced how \methodname builds a skill library from heterogeneous scientific resources, we next evaluate it from three complementary perspectives. First, we characterize the mined skill library through statistical analysis and internal validation, focusing on its validity, novelty, and overall composition. Second, we assess the utility of the skills already mined by \methodname on MoSciBench~\citep{liu2026towards}, asking whether the current library improves agent performance on scientific discovery tasks. Third, we examine whether \methodname can construct new skills for concrete user-specified scientific workflows on two challenging genomics tasks: cell type annotation for spatial transcriptomics data and identification of disease-relevant cells from single-cell RNA-seq data using the statistical genetics tool scDRS.

\subsection{Skill Library Overview}

We first summarize the mined library before turning to downstream evaluation. Because \methodname~is continuously expanding and self-evolving, the following statistics provide a snapshot at the time of writing. At this point, \methodname~contains 286 skills spanning 27 domains and 254 subdomains, mined from 394 resources (Fig.~\ref{fig:novelty_runtime}a). Every candidate skill is subjected to execution testing and synthetic-data testing, and skills whose intended use depends on cluster infrastructure are additionally evaluated with system testing. When a candidate skill fails these checks, the framework enters a repair and optimization loop and retests the skill until it either passes or is removed from consideration. As a result, all retained skills in the current library pass the required validation tests. To assess novelty, \methodname~searches existing skill libraries, such as SkillHub~\citep{skillhub_website} and SkillSMP~\citep{skillsmp_website}, using keywords derived from the candidate skill's task description. A mined skill is considered novel if no existing skill is found that completes the same task. Among the mined skills, 71.1\% satisfy this criterion and are added as new leaves in the domain knowledge tree. The remaining 28.9\% have related matches in existing libraries. When an existing skill has the same task description and method scope, the mined skill is discarded as redundant. When the existing and mined skills are complementary, the framework merges them and keeps the merged result as a leaf during tree refinement.

We next evaluate the runtime of skill discovery. Fig.~\ref{fig:novelty_runtime}b reports the average running time of the major stages under a fixed resource-mining budget. Skill extraction is the dominant cost, followed by resource mining, whereas testing and tree updates contribute relatively little overhead. We cap the search budget during resource mining to avoid unbounded exploration. The reported cost therefore reflects runtime rather than API token usage, and it measures controlled retrieval under a fixed search budget rather than open-ended search. 

Taken together, these results indicate that \methodname~can build a large and substantially novel skill library with moderate time cost. 
However, internal validation serves only as a screening step, not as final evidence of downstream usefulness. We therefore turn next to benchmark and workflow evaluations to assess whether the mined skills translate into practical gains on scientific tasks.

\begin{table}[t]
\centering
\resizebox{\columnwidth}{!}{
\begin{tabular}{lccccccccc}
\toprule[1.2pt]
& \multicolumn{3}{c}{health spa}
& \multicolumn{3}{c}{massspecgym}
& \multicolumn{3}{c}{pop genetics} \\
\cmidrule(lr){2-4} \cmidrule(lr){5-7} \cmidrule(lr){8-10}
Setting
& Repo-Acc (\%) & Paper-Acc (\%) & Exec (\%)
& Repo-Acc (\%) & Paper-Acc (\%) & Exec (\%)
& Repo-Acc (\%) & Paper-Acc (\%) & Exec (\%) \\
\midrule
w/o skills
& 64.71 & 47.06 & 100.00
& 40.00 & 40.00 & 100.00
& 53.85 & 30.77 & 100.00 \\
w/ skills
& \textbf{76.47} & \textbf{58.82} & 100.00
& \textbf{46.67} & \textbf{46.67} & 100.00
& \textbf{61.54} & \textbf{46.15} & 100.00 \\
\end{tabular}
}

\vspace{0.5em}

\resizebox{\columnwidth}{!}{
\begin{tabular}{lccccccccc}
\specialrule{0.8pt}{0pt}{0pt}
& \multicolumn{3}{c}{nurse stress}
& \multicolumn{3}{c}{cyclone}
& \multicolumn{3}{c}{terra} \\
\cmidrule(lr){2-4} \cmidrule(lr){5-7} \cmidrule(lr){8-10}
Setting
& Repo-Acc (\%) & Paper-Acc (\%) & Exec (\%)
& Repo-Acc (\%) & Paper-Acc (\%) & Exec (\%)
& Repo-Acc (\%) & Paper-Acc (\%) & Exec (\%) \\
\midrule
w/o skills
& 80.00 & 66.67 & 100.00
& 64.29 & 42.86 & 100.00
& 64.29 & 35.71 & 100.00 \\
w/ skills
& 80.00 & 66.67 & 100.00
&\textbf{ 71.43} & \textbf{57.14} & 100.00
& \textbf{64.29} & \textbf{42.86} & 100.00 \\
\bottomrule[1.2pt]
\end{tabular}
}
\vspace{-0.3em}
\caption{Results on all six datasets in MoSciBench for the same coding agent with and without \methodname~skills. Repo-Acc uses the scoring rule implemented in the official MoSciBench repository, whereas Paper-Acc uses the scoring rule described in the MoSciBench paper. Exec denotes code execution success rate. \methodname~improves performance on five of the six datasets and leaves one unchanged, while maintaining perfect execution success throughout.}
\label{tab:moscibench}
\vspace{-1em}
\end{table}

\subsection{MoSciBench}
In this section, we evaluate whether the skills mined by \methodname~improve performance on open-ended scientific tasks using MoSciBench~\citep{liu2026towards}, a benchmark for end-to-end multimodal scientific discovery with LLM agents. 
MoSciBench contains six datasets covering a broad range of scientific domains, including climate science, biomedical engineering, cheminformatics, health psychology, population genomics, and earth science.
We choose this benchmark because its tasks require multimodal scientific data integration, modeling, and hypothesis-driven reasoning, rather than isolated prediction problems within a single modality. Success depends not only on access to tools, but also on reusable procedural knowledge for coordinating evidence across modalities, conducting intermediate analyses, and sustaining multi-step scientific reasoning. We therefore compare the performance of the same coding agent with and without \methodname~skills across the full MoSciBench benchmark.

Table~\ref{tab:moscibench} presents the results. Adding \methodname~skills improves performance on five of the six datasets and leaves one unchanged. Averaged over all six datasets, Repo-Acc increases from 61.19\% to 66.73\%, and Paper-Acc increases from 43.85\% to 53.05\%. The improvement is most evident on \textit{health spa}, \textit{pop genetics}, and \textit{cyclone}. The skill-augmented agent also improves on \textit{massspecgym} and \textit{terra}, while \textit{nurse stress} remains unchanged, likely because the baseline is already strong on this dataset. At the same time, the code execution success rate remains 100.00\% on all six datasets in both settings. This suggests that the gains come from better task completion and scientific reasoning rather than from easier execution alone. Overall, these results show that the mined library provides useful procedural guidance that improves benchmark performance while preserving stable execution.

\subsection{Cell Type Annotation}
After showing that existing skills in \methodname~improve agent performance on the MoSciBench benchmark, we conduct an in-depth analysis on two challenging genomics tasks, cell type annotation and scDRS. We show that \methodname~can construct new skills for concrete user-specified scientific workflows and improve the performance of a coding agent on these domain-specific tasks.

The first task is cell type annotation for spatial transcriptomics data. The goal is to assign each cell to a biologically meaningful identity using molecular profiles and spatial context, which is foundational for many downstream analyses~\citep{shen2022spatial}. This task requires multi-step procedural execution including preprocessing, representation learning, marker or reference reasoning, and result validation, making it a suitable setting to test whether agents can reliably execute workflows rather than just call a single tool. 

We use a developing human heart MERFISH dataset~\citep{farah2024spatially} containing 228,635 cells, 238 genes, and 3 samples. For this task, rather than selecting from a fixed skill library, we provide \methodname~with a task prompt and allow it to retrieve relevant resources, construct candidate skills, and test them for the task. We then evaluate Codex with and without the resulting skill, as well as SpatialAgent~\citep{wang2025spatialagent}, a domain-specific agent for spatial transcriptomics analysis. To ensure a fair comparison, we provide the same dataset and prompt to all methods and harmonize both ground-truth labels and method outputs into eight major cell types following the SpatialAgent paper~\citep{wang2025spatialagent}.

For this challenging task, different methods adopt distinct workflows. Vanilla Codex uses PCA followed by k-means clustering, with marker-set scoring and ambiguity handling for cell type assignment. 
Codex with the \methodname-generated skill instead applies a marker-signature-based pipeline that integrates PCA, neighborhood graph construction with Leiden clustering, centroid-similarity-based provisional labeling, cluster-level rule refinement, and local neighborhood validation. SpatialAgent uses a domain-specific workflow with expert-curated tools: it first retrieves an external single-cell RNA-seq reference dataset from the CZI database, transfers cell type labels using Harmony, and then annotates clusters based on the transferred labels and marker genes.

\begin{figure}[t]
\centering
\includegraphics[width=\linewidth]{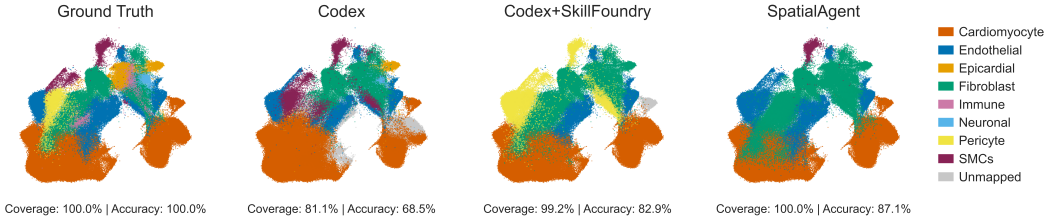}
\caption{UMAP of cell-type annotations from ground truth, Codex (without skill), Codex+\methodname, and SpatialAgent. 
SMCs denote smooth muscle cells.
Labels are harmonized into eight major cell types, following the SpatialAgent~\citep{wang2025spatialagent} paper. Coverage is the fraction of cells assigned to one of the eight major cell types, and accuracy is the fraction whose predicted label matches the harmonized ground truth.}
\label{fig:cell_type_annotation}
\vspace{-1.3em}
\end{figure}

Quantitatively, Codex+\methodname markedly outperforms vanilla Codex in both coverage and accuracy, reaching 99.2\% coverage and 82.9\% accuracy, compared with 81.1\% and 68.5\%, respectively (Figure~\ref{fig:cell_type_annotation}). In this setting, higher coverage indicates that the agent is able to assign labels to a larger proportion of cells rather than leaving them unresolved, whereas higher accuracy indicates that those assigned labels better match the ground truth. SpatialAgent performs best overall, achieving 
100.0\% coverage and 87.1\% accuracy. This stronger performance is expected, as SpatialAgent leverages an external single-cell reference dataset, whereas Codex does not rely on external reference data or services. Overall, these results show that \methodname can successfully construct new skills for user-specified scientific workflows, and that adding the resulting skill substantially improves Codex on this challenging cell type annotation task.

\subsection{scDRS Workflow}
Besides the cell type annotation task, we evaluate task-specific skill design on the scDRS~\citep{zhang2022polygenic} workflow. scDRS links GWAS summary statistics with single-cell RNA-seq data to identify disease-relevant cells or cell types, and therefore requires the agent to coordinate heterogeneous inputs, manage dependencies and formats, follow a multi-step statistical procedure, and produce interpretable outputs. 
Compared with benchmark-style tasks, it is a stronger test of workflow competence. 
In this setting, we ask whether \methodname~can design a useful skill for the scDRS workflow from a user request, and whether the resulting skill can improve the performance of an external scientific agent such as Biomni~\citep{huang2025biomni}. 
Biomni is a general-purpose biomedical agent with a large tool and database ecosystem, but scDRS is not part of its existing knowledge base or predefined workflows. 
We therefore compare Biomni with and without the \methodname~skill,
packaging the generated skill directly as an external resource to keep the comparison minimally invasive.

We evaluate transfer to an external scientific agent by comparing Biomni with and without these skills on the scDRS workflow using the TMS FACS \citep{tabula2020single} scRNA-seq dataset together with height GWAS. 
We run each setting for three replicates and compare the outputs against expert reference analyses in a blinded evaluation by statistical genetics experts. The evaluation includes both qualitative and quantitative criteria. The qualitative score measures whether a run returns the key expected analyses and artifacts, including individual cell, cell-type, and gene-level results, proper FDR correction, identification of chondrocytes as the top associated cell type \citep{baron2015short}, within-cell-type heterogeneity analysis, and all relevant output files. The quantitative score is the RMSE between the cell-level disease scores produced by each run and the expert reference results.

\begin{figure}[t]
\centering
\includegraphics[width=0.85\linewidth]{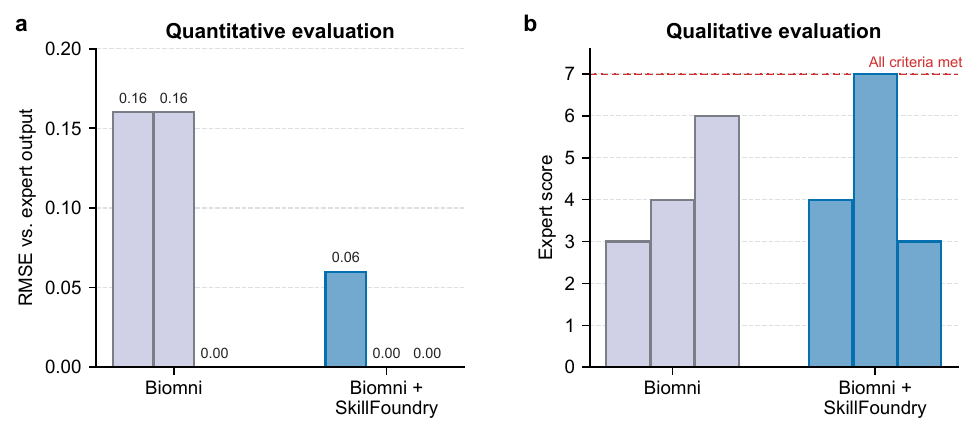}
\vspace{-0.9em}
\caption{Quantitative and qualitative evaluation of Biomni on the scDRS workflow, with and without \methodname skills. 
(a) RMSE between agent and expert outputs across three replicates. 
\methodname reduces error overall, with two runs exactly matching the expert output.
(b) Expert qualitative scores across the same replicates.
A score of 7 indicates that all evaluation criteria are met.
\methodname produces the only run that satisfies all qualitative criteria.}
\label{fig:scdrs}
\vspace{-1em}
\end{figure}

Results are shown in Fig.~\ref{fig:scdrs}. Qualitatively, one Biomni+\methodname~run is the only setting that satisfies all evaluation criteria (Fig.~\ref{fig:scdrs}b). In addition, Biomni+\methodname~produces more meaningful and interpretable figures than vanilla Biomni (Fig.~\ref{app:fig-biomni-wo-skill} and Fig.~\ref{app:fig-biomni-w-skill}; Section~\ref{app:scdrs_details}). Quantitatively, it is also more accurate and more robust. Two of the three skill-augmented runs exactly match the expert runs, and the mean RMSE between the workflow output scores and the expert-run scores is substantially lower than that of the baseline, decreasing from 0.11 to 0.02 (Fig.~\ref{fig:scdrs}a). By contrast, Biomni without skills often omits the \texttt{filter-data} parameter, which causes scDRS to score unfiltered noisy genes and leads to incomplete or inaccurate outputs. 
Further comparison of representative analysis figures produced in the two settings is shown in App.~\ref{app:scdrs_details}.
Without the synthesized skills, Biomni generates only a basic ranking plot based on nominal significance. With \methodname~skills, it produces a richer and more accurate summary that combines association ranking with FDR-based support and within-cell-type heterogeneity information. Together, these results show that \methodname~improves not only workflow execution, but also the quality and interpretability of the resulting scientific artifacts.

Overall, the experiments support two complementary claims. First, \methodname~can automatically build a valid and substantially novel scientific skill library from heterogeneous resources. Second, the same framework can design task-specific skills for concrete scientific objectives. The gains on MoSciBench, cell type annotation, and scDRS suggest that the resulting skills are not only internally valid, but also useful in realistic benchmark and scientific workflow settings.
\subsection{Limitations and Future Work}
We acknowledge several limitations in this study. 
First, the current coverage of \methodname~is still relatively limited at the time of writing. 
Although the library is intended to expand continuously as new resources are mined, validated, and refined, the present snapshot still covers only a subset of scientific domains and task families. 
Second, most mined skills have so far been validated primarily through the framework’s internal testing pipeline rather than through broad downstream evaluation. 
Although these internal tests are useful for ensuring skill executability, they do not fully capture practical utility or failure modes across real scientific workflows. 
Third, our current evaluation remains limited in scope and is concentrated on a small number of domains and tasks. The current skill library should therefore be used with appropriate caution, especially in settings that require high reliability. 
Future work should broaden domain coverage and strengthen expert-informed downstream evaluation of mined skills across more realistic workflows, while also improving long-term maintenance and novelty assessment as the library grows.

\section{Conclusion}
Modern scientific ecosystems already contain abundant procedural knowledge across repositories, APIs, scripts, notebooks, documentation, databases, and papers, but much of this knowledge remains fragmented across heterogeneous resources that agents cannot directly use. \methodname addresses this gap by converting such resources into structured agent skills through a tree-guided closed loop for mining, validation, and refinement. 
Empirically, the resulting skill library is substantially novel and internally valid.
We demonstrate that \methodname improves agent performance on both standardized benchmarks and real-world scientific workflows. 
More broadly, \methodname points to a path for transforming fragmented scientific know-how into reusable capabilities and, in turn, making scientific agents more capable, scalable, and reliable.

\bibliography{colm2026_conference}
\bibliographystyle{colm2026_conference}

\newpage
\appendix
\newpage
\section{Appendix}

\subsection{Method}
\label{app:method}

\subsubsection{Tree State and Branch Prioritization}
Beyond its role as a topical taxonomy, the domain knowledge tree serves as a stateful controller for skill acquisition. Each node stores lightweight repository state, including linked resources, linked skills, recent verification outcomes, and a coarse coverage flag. These signals are maintained through machine readable registries for taxonomy structure, provenance, and skill metadata. As a result, the tree functions not only as an ontology over the target domain, but also as an index of the current library state.

This state is used to prioritize where the next mining cycle should focus. In particular, \methodname~does not choose resources independently. It first identifies branches or leaves whose expected marginal value is high under the current repository state. Examples include leaves that already have useful resources but no executable skills, leaves supported only by low confidence starter skills, and leaves whose current skills repeatedly fail validation. This branch first prioritization reduces redundant search and turns a large heterogeneous resource space into a sequence of bounded skill acquisition problems.

The initial taxonomy is manually curated but intentionally broad. It covers computational biology together with adjacent scientific software areas such as chemistry, visualization, workflows, spatial transcriptomic analysis, and scientific agents. The tree is then revised over time. Broad branches may be split once mining reveals stable subareas, while stale or weakly supported branches may be merged, collapsed, or marked for lower priority. In this sense, the tree is neither fixed nor purely data driven. It is a structured prior that is updated according to observed resource coverage and skill outcomes.

\subsubsection{Skill Packaging and Repository Integration}

After a branch is selected, \methodname~retrieves a branch-conditioned pool of candidate resources rather than performing an unrestricted crawl. From these resources, the system extracts an operational contract for each candidate skill. This contract specifies the task scope of the capability, expected inputs and outputs, environment and dependency assumptions, execution steps, provenance links, follow up guidance, and, when available, example invocations or test commands.

The extracted contract is then compiled into a concrete skill package. In the implementation, a compiled skill is more than a text description. It contains a human readable skill specification, machine readable metadata, provenance records, executable scripts, example assets, and repository level tests whenever such artifacts can be resolved. This packaging choice is important because it turns induced skills into reusable software objects that remain interpretable to agents while still being maintainable at the repository level.

Whenever possible, each candidate skill is paired with a smoke target or test command before integration. The resulting status is written to the shared skill registry so that later cycles can distinguish between untested, repaired, and verified entries. The same contract format is also reused in prompt driven skill design for user specified tasks. This means that task specific skill synthesis does not introduce a separate pathway for ad hoc tool creation, but instead reuses the same packaging and validation interface as the main mining loop.

\subsubsection{Hierarchical Validation, Repair, and Optimization}

The self improving behavior of \methodname~comes from a hierarchical validation process that goes beyond simple execution checks. The first layer is correctness oriented execution testing. The system runs a skill through its resolved smoke target or registered test command and checks whether the declared contract is actually executable. When this layer fails, the framework enters a repair loop that attempts to diagnose the failure, revise the implementation or test surface, and rerun the check. This establishes a minimum executability requirement before a skill can be considered for promotion.

The second layer asks whether the skill is useful as an agent abstraction. A skill may run successfully while still being too incomplete, too brittle, or too poorly structured to help an agent. To address this, \methodname~constructs skill versus no skill comparisons on task specific cases and checks whether the maintained wrapper provides a meaningful advantage over an ad hoc baseline. If the advantage is weak or inconsistent, the framework enters an optimization loop that revises the skill and benchmarks it again.

In addition, the framework uses two complementary validation surfaces that are especially important in scientific settings. System testing is used for skills whose realistic deployment depends on infrastructure such as SLURM, environment modules, or cluster-specific execution conventions. Synthetic data testing is used when live resources are unavailable, unstable, rate limited, or too expensive for routine regression testing. Rather than approximating the full downstream task, these tests verify whether the skill handles controlled inputs correctly, respects expected file and argument contracts, and produces stable outputs under deterministic mock conditions. This makes them useful for catching interface errors, broken assumptions, and incomplete packaging even when real world execution is not practical.

\subsubsection{Novelty Review and Tree Update}

Passing validation is still not sufficient for acceptance into the library. A candidate skill may be correct and useful while contributing little marginal value if it largely duplicates an existing abstraction. \methodname~therefore performs a final novelty and redundancy review against both the local library and external skill ecosystems. This review is not based only on lexical similarity. It also considers topic placement, provenance, intended scope, and whether the new skill adds genuinely different capability, coverage, or usability.

Depending on the outcome, a candidate skill may be accepted as a new leaf, merged into an existing entry, marked for review, or deprioritized in future updates. These decisions are then written back to the domain tree. Skills that pass become new or improved leaves. Skills that repeatedly fail leave a branch marked as only partially covered. Skills that are found to be redundant or low value may trigger consolidation or pruning of the corresponding branch. In this way, validation outcomes do not merely filter skills locally, but actively reshape the search structure used in subsequent mining rounds.

\subsubsection{Parallel Acquisition and Controlled State Mutation}

Although branch level exploration can be parallelized, \methodname~is conservative in how it mutates shared repository state. Resource search and skill construction can run in isolated worker workspaces, but only leaf owned artifacts are merged automatically. Shared global files, including registries, reports, and site summaries, remain refresh owned. This separation prevents concurrent workers from writing inconsistent global state and ensures that each mining cycle begins from a coherent repository snapshot.

At a larger timescale, the framework supports checkpointable campaigns that alternate branch focused mining with batched evaluation over pending skills. This allows the library to grow incrementally while preserving auditability. More broadly, it reflects a central design principle of \methodname. Exploration may be parallel, but integration should remain explicit, deterministic, and easy to inspect.

\subsubsection{Implementation Details}

We instantiate \methodname~as a staged pipeline whose internal modules align with the phases in Fig.~\ref{Fig:overview}. Specifically, \texttt{tree\_check} corresponds to \emph{Domain Tree}, \texttt{resource\_search} to \emph{Resource Mining}, \texttt{skill\_build} together with task-specific \texttt{design\_skill} to \emph{Skill Extraction}, \texttt{skill\_test}, \texttt{layer1\_fix}, \texttt{layer2\_benchmark}, and \texttt{layer2\_optimize} to \emph{Skill Testing}, and \texttt{refresh} plus \texttt{novelty\_check} to the post-validation tree update supporting \emph{Tree Expansion} and \emph{Tree Refinement}.

For model allocation, we use GPT-5.4 with medium reasoning effort for \texttt{tree\_check}, \texttt{skill\_build}, \texttt{refresh}, and \texttt{design\_skill}; GPT-5.4 with high reasoning effort for \texttt{resource\_search}; and GPT-5.4-mini with medium reasoning effort for \texttt{skill\_test}, \texttt{layer1\_fix}, \texttt{layer2\_benchmark}, \texttt{layer2\_optimize}, and \texttt{novelty\_check}. 
This allocation makes \texttt{resource\_search} the only default high-effort stage in the main loop, reflecting that searching and triaging heterogeneous scientific resources is the most reasoning-intensive step. By contrast, skill construction remains on the larger model at medium effort, while validation, repair, benchmarking, and novelty review are delegated to the smaller model for efficiency.

For downstream evaluation, we use Codex with {GPT-5.1-Codex-mini} and the default medium reasoning effort on MoSciBench, and GPT-codex-5.3 with medium reasoning effort for cell type annotation. For the scDRS transfer experiment, \methodname~first synthesizes task-specific skills and then exports them to Biomni as external skill resources. Within each setting, we keep the underlying agent configuration fixed and vary only the availability of the relevant \methodname~skills. Full prompting templates, model versions, run counts, and hardware details are deferred to the appendix.

\subsubsection{Prompt Templates}
We use stage-conditioned prompt templates rather than a single monolithic instruction. A cycle controller governs iterative repository mining, specialized prompts handle isolated leaf workers and user-specified skill design, and a hierarchical evaluation suite repairs failures, benchmarks skill utility, and checks novelty.

We show abbreviated prompt templates rather than all instantiated prompts. Repeated fields such as repository summaries, artifact directories, and strict JSON response schemas are compressed for readability.

\begin{promptblock}
\small
\textbf{\textcolor{PromptBlue}{System Prompt:}} You are operating inside the SkillFoundry repository. Follow the workflow contract in \texttt{experiments.md} and reuse existing repository scripts instead of creating parallel registries.

\textbf{\textcolor{PromptRed}{Prompt Inputs:}} \texttt{STAGE=\{stage\}}, \texttt{MODE=cycle}, repository summary, focus leaves, and a stage-specific objective.

\textbf{Key Constraints:} Work directly in the current checkout when appropriate; reuse repository scripts and registries; keep verification labels honest.

\textbf{Response Contract:} Return only a JSON object matching the stage schema.
\end{promptblock}

\vspace{0.5em}

\begin{promptblock}
\small
\textbf{\textcolor{PromptBlue}{System Prompt:}} You are operating inside an isolated workspace copy of the SkillFoundry repository. This is a parallel worker stage for a single taxonomy leaf.

\textbf{\textcolor{PromptRed}{Prompt Inputs:}} \texttt{STAGE=\{stage\}}, \texttt{MODE=parallel\_leaf\_stage}, repository summary, target leaf, stage objective, and artifact directory.

\textbf{Key Constraints:} Stay scoped to the assigned leaf; prefer edits under skill-local files and leaf-specific tests; do not modify shared global files such as registries, site files, \texttt{README.md}, or planning documents.

\textbf{Response Contract:} Return only a JSON object, and record any shared-file follow-up in structured fields such as \texttt{repo\_changes}, \texttt{blockers}, or \texttt{next\_steps}.
\end{promptblock}

\vspace{0.5em}

\begin{promptblock}
\small
\textbf{\textcolor{PromptBlue}{System Prompt:}} You are operating inside the SkillFoundry repository. This mode is for a user-specified task: search resources first when the request needs additional materials, then implement and test the resulting skill.

\textbf{\textcolor{PromptRed}{Prompt Inputs:}} \texttt{STAGE=design\_skill}, \texttt{MODE=design\_skill}, user task prompt, repository summary, and focus leaves.

\textbf{Key Constraints:} Search for canonical papers, repositories, notebooks, package references, or workflows when needed; build or refine one concrete skill path; add or update repository-level tests and, when relevant, a Slurm execution path.

\textbf{Response Contract:} Return only a JSON object matching the design-skill schema.
\end{promptblock}

\vspace{0.5em}

\begin{promptblock}
\small
\textbf{\textcolor{PromptBlue}{System Prompt:}} Hierarchical evaluation operates in multiple modes: failure repair, task-specific benchmarking, benchmark-driven optimization, and novelty checking.

\textbf{\textcolor{PromptRed}{Prompt Inputs:}} Skill metadata, repository summary, failure records or benchmark results, local similarity candidates, and an artifact directory.

\textbf{Key Constraints:} Diagnose failures before editing; apply targeted fixes rather than broad refactors; benchmark with-skill versus no-skill baselines conservatively; optimize only when benchmark deficits are clear; and assess overlap against both local skills and external ecosystems.

\textbf{Response Contract:} Return structured JSON containing repair findings, benchmark scores, optimization actions, or novelty scores, depending on the evaluation sub-stage.
\end{promptblock}

\vspace{0.5em}

\begin{promptblock}
\small
\textbf{\textcolor{PromptBlue}{System Prompt:}} Design or refine an experiment-only portable skill for a specific task.

\textbf{\textcolor{PromptRed}{Prompt Inputs:}} Task title, task goal, target directory, preferred methods, family-specific starter guidance, and required deliverables.

\textbf{Key Constraints:} Modify only files under the task-specific target directory; do not edit repository-level registries, shared skill libraries, site files, \texttt{README.md}, \texttt{experiments.md}, or planning files.

\textbf{Response Contract:} Produce only the required experiment-local deliverables for the designated task.
\end{promptblock}

\subsection{Experiments}

\subsubsection{scDRS}

\label{app:scdrs_details}

Here we provide the evaluation criteria, the replicate-level outcomes, and representative analysis figures produced by Biomni with and without the task-specific skills synthesized by \methodname.

We compare Biomni with and without \methodname~skills on the scDRS workflow using the TMS FACS \citep{tabula2020single} scRNA-seq dataset together with height GWAS. Each setting is run three times and compared against an expert reference analysis. We evaluate the outputs from two complementary perspectives. The qualitative score ranges from 0 to 7, with one point assigned for each of the following criteria: returning individual cell-level, cell-type-level, and gene-level analyses, applying proper FDR correction, identifying chondrocytes as the top associated cell type \citep{baron2015short}, performing within-cell-type heterogeneity analysis, and returning useful output files. The quantitative metric is the RMSE between the cell-level disease scores produced by each run and the expert reference results. We report both metrics because numerical agreement alone does not guarantee that the resulting scientific artifact is complete and interpretable.

\begin{table}[!ht]
\centering
\small

\resizebox{\columnwidth}{!}{
\begin{tabular}{ccccc}
\toprule
Replicate & Setting & Qualitative score & RMSE & Notes \\
\midrule
1 & Biomni & 3 & 0.16 & Filtering option not enabled \\
2 & Biomni & 4 & 0.16 & Minor deviation from expert pipeline \\
3 & Biomni & 4 & 0.00 & Exact scores but incomplete analysis \\
4 & Biomni + \methodname & 6 & 0.06 & Nearly complete analysis \\
5 & Biomni + \methodname & 7 & 0.00 & All criteria met, exact numerical match \\
6 & Biomni + \methodname & 3 & 0.00 & Exact scores but incomplete reporting \\
\bottomrule
\end{tabular}
}
\caption{Replicate-level scDRS evaluation. Replicates 1 to 3 correspond to Biomni without \methodname~skills, and replicates 4 to 6 correspond to Biomni with \methodname~skills.}
\label{tab:scdrs_replicates}
\end{table}

Table.~\ref{tab:scdrs_replicates} shows that \methodname~improves both accuracy and robustness. Biomni+\methodname~is the only setting that satisfies all seven qualitative criteria, and it also produces exact agreement with the expert scores in two of three runs, compared with one of three runs for Biomni alone. Its mean RMSE is substantially lower, decreasing from 0.11 to 0.02. The replicate-level breakdown also shows why both metrics are needed. Some runs recover the correct cell-level scores but still fail to return the full set of analyses or outputs needed for scientific interpretation.

\begin{figure}[!ht]
    \centering
    \includegraphics[width=0.7\linewidth]{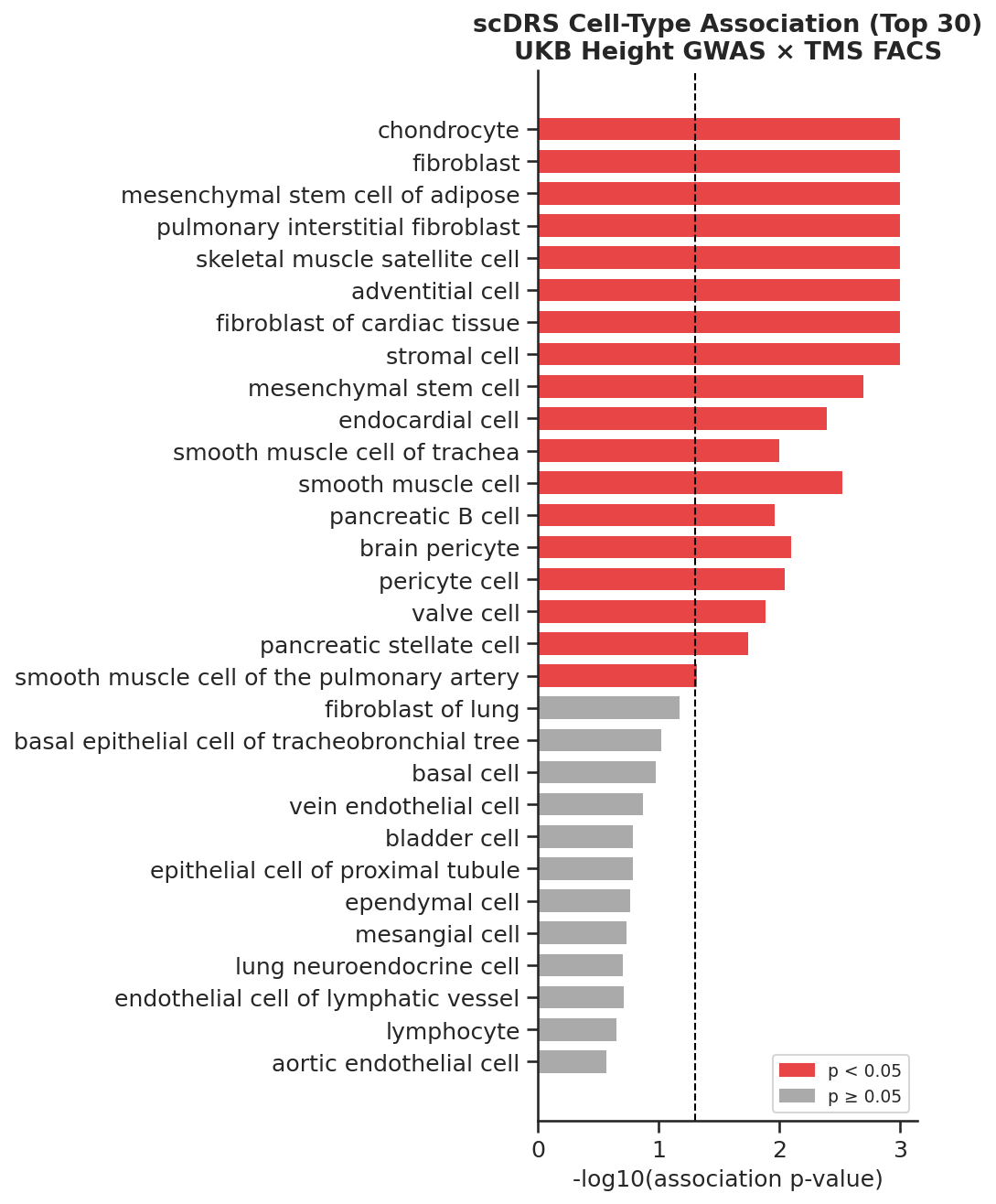}
    \caption{Representative scDRS output generated by Biomni without \methodname~skills. The figure ranks cell types by association strength and marks nominal significance, but does not summarize FDR-supported signal or within-cell-type heterogeneity.}
    \label{app:fig-biomni-wo-skill}
\end{figure}

\begin{figure}[!ht]
    \centering
    \includegraphics[width=0.85\linewidth]{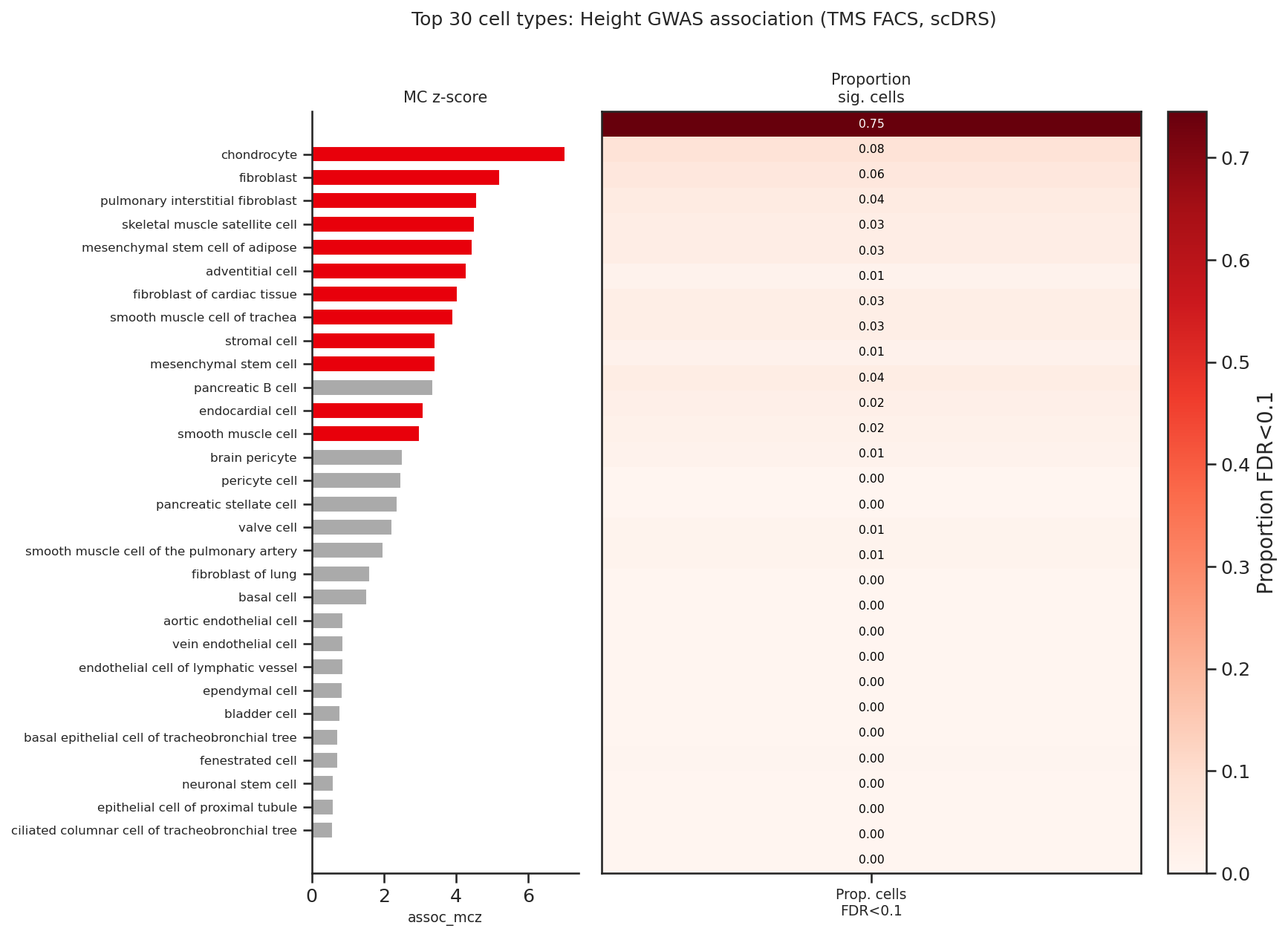}
    \caption{Representative scDRS output generated by Biomni with \methodname~skills. In addition to ranking cell types by association strength, the figure reports the proportion of FDR-significant cells within each cell type, providing a richer summary of statistical support and cell-type heterogeneity.}
    \label{app:fig-biomni-w-skill}
\end{figure}

Fig.~\ref{app:fig-biomni-wo-skill} and Fig.~\ref{app:fig-biomni-w-skill} compare representative analysis figures produced in the two settings. Without the synthesized skills, Biomni generates only a basic ranking plot based on nominal significance. This is enough to show a rough ordering of top cell types, but it does not summarize FDR-supported signal or within-cell-type heterogeneity. With \methodname~skills, Biomni produces a richer and more interpretable figure that combines ranked cell-type associations with an additional panel showing the proportion of FDR-significant cells in each cell type. As a result, the with-skill figure provides a more complete summary of both statistical support and cell-type heterogeneity, which helps explain the higher qualitative scores.

\begin{figure}[!ht]
    \centering
    \includegraphics[width=0.9\linewidth]{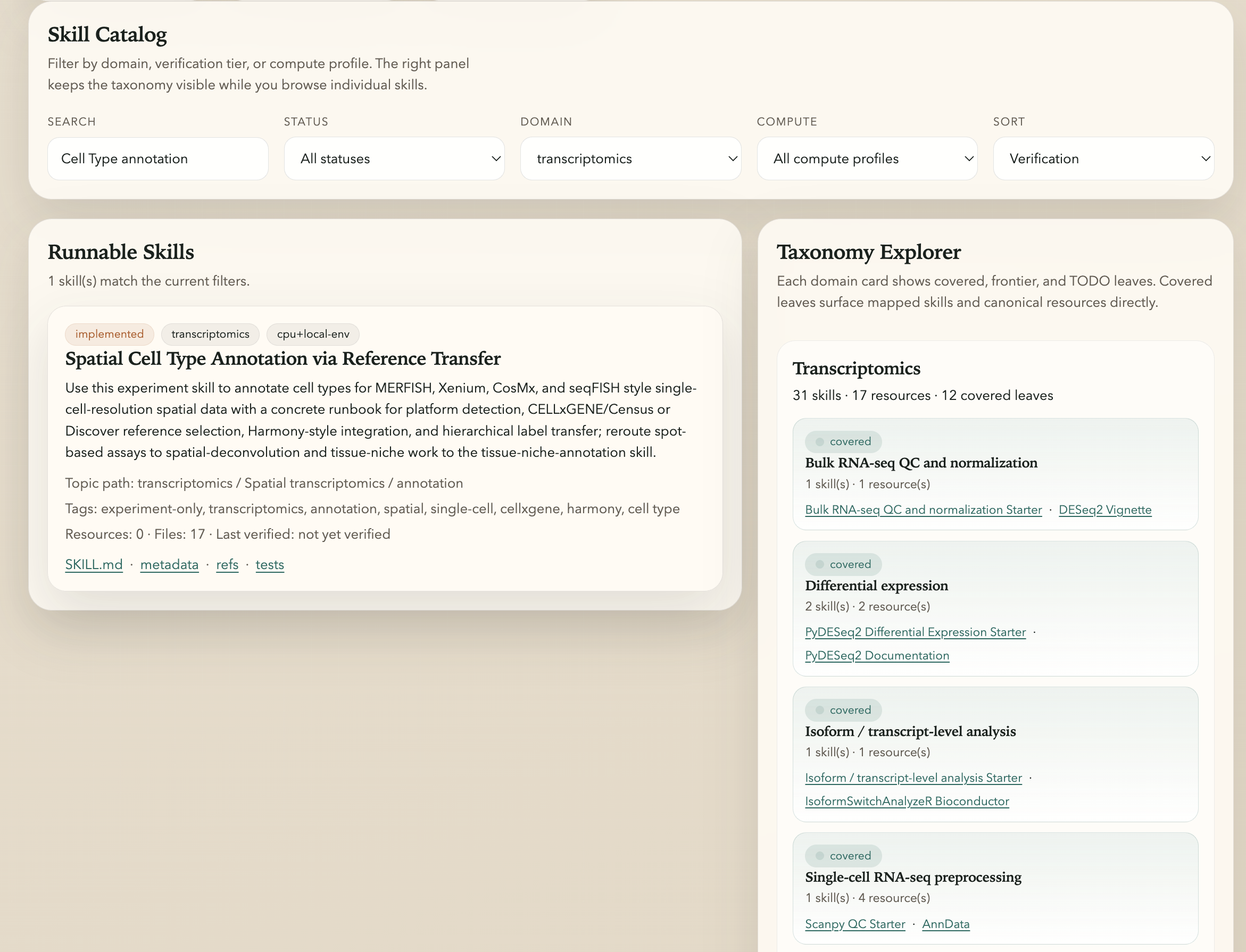}
    \caption{Web dashboard view of a cell type annotation skill. The interface shows searchable skill metadata, taxonomy placement, and links to associated artifacts and source resources.}
    \label{app:screenshot-annotation}
\end{figure}

\begin{figure}[!ht]
    \centering
    \includegraphics[width=0.9\linewidth]{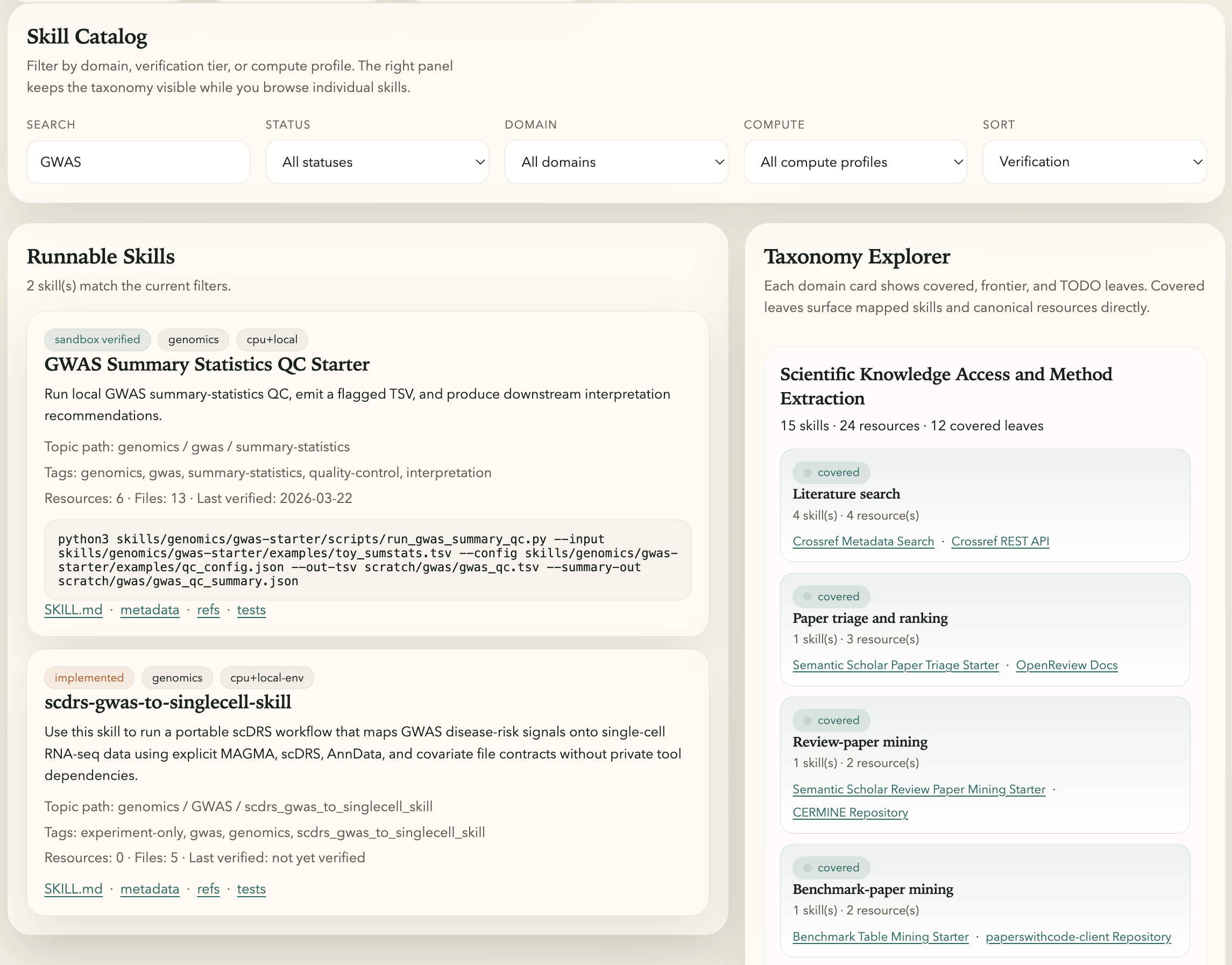}
    \caption{Web dashboard view of scDRS-related skills. The interface supports search and filtering over runnable skills while exposing their domain assignments, verification status, and provenance information.}
    \label{app:screenshot-gwas}
\end{figure}

\subsubsection{Dashboard}
We also provide a web dashboard for browsing, monitoring, and managing the mined skill library. The dashboard presents each skill together with its verification status, domain assignment, compute profile, and other metadata, and supports interactive search and filtering across the library. A taxonomy explorer further situates each skill within the domain tree and summarizes coverage at different levels of granularity. In addition, the interface exposes provenance information, allowing users to inspect the resources from which a skill was mined and to access associated artifacts such as the skill specification, metadata, references, and tests. Fig.~\ref{app:screenshot-annotation} and Fig.~\ref{app:screenshot-gwas} show example views for a cell type annotation skill and scDRS-related skills, respectively.


\end{document}